\pdfoutput=1
\documentclass[11pt,a4paper]{article}
\usepackage[hyperref]{AACL-IJCNLP2020}
\usepackage{times}
\usepackage{latexsym}

\usepackage{microtype}

\aclfinalcopy % Uncomment this line for the final submission
\usepackage{booktabs}
\usepackage{amssymb}
\usepackage{pifont}
\usepackage{multirow}
\usepackage{graphicx}

\usepackage{color}

\newcommand{\fairseq}{\textsc{fairseq}}
\newcommand{\stot}{\textsc{fairseq S2T}}

\title{\stot: Fast Speech-to-Text Modeling with \fairseq}

\author{Changhan Wang$^1$, Yun Tang$^1$, Xutai Ma$^{1,2}$, Anne Wu$^1$, Sravya Popuri$^1$, \\\textbf{Dmytro Okhonko$^1$, Juan Pino$^1$} \vspace*{0.2cm} \\
$^1$Meta - Fundamental AI Research (FAIR) \\
$^2$Johns Hopkins University \vspace*{0.2cm} \\
  {\tt xutai\_\thinspace ma@jhu.edu} \\
\texttt{\{changhan,yuntang,annewu,spopuri,oxo,juancarabina\}@fb.com} \\
}

\date{}

\begin{document}
\maketitle
\begin{abstract}
We introduce \stot, a \fairseq~\citep{ott2019fairseq} extension for speech-to-text (S2T) modeling tasks such as end-to-end speech recognition and speech-to-text translation. It follows \fairseq's careful design for scalability and extensibility. We provide end-to-end workflows from data pre-processing, model training to offline (online) inference. We implement state-of-the-art RNN-based, Transformer-based as well as Conformer-based models and open-source detailed training recipes. \fairseq's machine translation models and language models can be seamlessly integrated into S2T workflows for multi-task learning or transfer learning. \stot~documentation and examples are available at \url{https://github.com/pytorch/fairseq/tree/master/examples/speech_to_text}.
\end{abstract}

\section{Introduction}

End-to-end sequence-to-sequence (S2S) modeling has witnessed rapidly increased applications in speech-to-text (S2T) tasks. It achieves state-of-the-art performance on automatic speech recognition (ASR)~\citep{park2019specaugment,synnaeve2019end} and leads to the recent resurgence of speech-to-text translation (ST) research~\citep{duong2016attentional,berard2016listen}. ASR and ST are closely related. There are recent attempts to combine the two tasks under the same S2S model architecture via multi-task learning~\citep{anastasopoulos2018tied,liu2019synchronous}. They also benefit from each other via transfer learning~\citep{bansal2018pre,wang2020improving} and are able to leverage additional supervision from machine translation (MT) and language modeling (LM). When supervised data is not abundant, self-supervised pre-training~\citep{Schneider2019,wu2020selfsupervised} and semi-supervised training~\citep{Kahn2020SelfTrainingFE,pino2020selftraining} lowers the requirements on supervision and improves model performance.

\begin{table*}[ht]
    \centering
    \small
    \begin{tabular}{c|ccccccc|cc}
    \toprule
     & \multirow{2}{*}{ASR} & \multirow{2}{*}{LM} & \multirow{2}{*}{MT} & Non-Autoreg. & Offline & Online & Speech & Multi-node & Pre-trained  \\
     & & & & MT & ST & ST & Pre-training & training & models \\
    \midrule 
    ESPNet-ST & \checkmark & \checkmark & \checkmark & & \checkmark & & & \checkmark$^\dagger$ & \checkmark \\
    Lingvo & \checkmark & \checkmark & \checkmark & & \checkmark$^\ddagger$ & & & \checkmark & \\
    OpenSeq2seq$^1$ & \checkmark & \checkmark & \checkmark & & & & & \checkmark & \checkmark \\
    RETURNN$^2$ & \checkmark & \checkmark & \checkmark & & \checkmark & & & \checkmark & \checkmark \\
    SLT.KIT$^3$ & \checkmark & & \checkmark & & \checkmark & & & & \checkmark \\
    Tensor2Tensor$^4$ & \checkmark & \checkmark & \checkmark & & & & & \checkmark & \checkmark \\
    OpenNMT$^5$ & \checkmark & \checkmark & \checkmark & & & & & \checkmark & \checkmark \\
    Kaldi$^6$ & \checkmark & \checkmark & & & & & & & \checkmark \\
    Wav2letter++$^7$ & \checkmark & \checkmark & & & & & & & \checkmark \\
    \midrule
    \textbf{fairseq S2T} & \checkmark & \checkmark & \checkmark & \checkmark & \checkmark & \checkmark & \checkmark & \checkmark & \checkmark \\
    \bottomrule
    \end{tabular}
\caption{Comparison of \stot~with counterpart toolkits (as of July 2020). $^\dagger$ Only available in version 2 (under development). $^\ddagger$ Not publicly available. $^1$ \citet{kuchaiev2018openseq2seq}. $^2$ \citet{zeyer-etal-2018-returnn}. $^3$ \citet{zenkel2018open}. $^4$ \citet{vaswani2018tensor2tensor}. $^5$ \citet{klein2017opennmt}. $^6$ \citet{povey2011kaldi}. $^7$ \citet{pratap2018w2l}.}
    \label{tab:counterparts}
\end{table*}

The increased connections among ASR, ST, MT and LM has called for all-in-one S2S modeling toolkits, and the use of large-scale unlabeled speech data sets the scalability requirements. In this paper, we introduce \stot, a \fairseq~\citep{ott2019fairseq} extension for S2T tasks such as end-to-end ASR and ST. It follows \fairseq's careful design for scalability and extensibility. We provide end-to-end workflows from data pre-processing, model training to offline (online) inference. We implement state-of-the-art RNN-based~\citep{chan2016listen,berard2018end}, Transformer-based~\citep{vaswani2017attention,Mohamed2019TransformersWC} and Conformer-based~\citep{gulati20_interspeech} models and open-source detailed training recipes. \fairseq's MT models and LMs can be seamlessly integrated into S2T workflows for multi-task learning or transfer learning. To facilitate model evaluation, we add a collection of scorers as well as VizSeq~\citep{wang-etal-2019-vizseq} integration for visualized error analysis. \stot~documentation and examples are available at \url{https://github.com/pytorch/fairseq/tree/master/examples/speech_to_text}.

With counterpart toolkits such as ESPNet~\citep{inaguma-etal-2020-espnet} and Lingvo~\citep{shen2019lingvo}, \stot~pursues the best integration, scalability and reproducibility. A detailed comparison of \stot~with its counterparts can be found in Table~\ref{tab:counterparts}.

\section{Features}

\paragraph{Fairseq Models} \fairseq~provides a collection of MT models~\citep{ng-etal-2019-facebook,lewis2019bart} and LMs~\citep{liu2019roberta,conneau2019unsupervised} that demonstrate state-of-the-art performance on standard benchmarks. They are open-sourced with pre-trained models. \fairseq~also supports other tasks such as text summarization, story generation and self-supervised speech pre-training.

\paragraph{S2T extension} \stot~adds attention-based RNN models~\citep{chan2016listen,berard2018end}, Transformer models~\citep{vaswani2017attention,Mohamed2019TransformersWC} as well as the latest Conformer models~\citep{gulati20_interspeech} for ASR and ST. It also supports CTC criterion~\citep{graves2006connectionist} for ASR. For the simultaneous ST setting, it includes online models with widely used policies: monotonic attention~\citep{raffel2017online}, wait-$k$~\citep{Ma2018STACLST},
monotonic infinite lookback attention~\citep{arivazhagan2019monotonic},
and monotonic multihead attention~\citep{ma2019monotonic}.

\paragraph{Data Pre-Processing}
\stot~extracts Kaldi-compliant~\citep{povey2011kaldi} speech features (e.g. log mel-filter banks) automatically from WAV/FLAC audio files via PyKaldi~\citep{pykaldi} or torchaudio\footnote{https://github.com/pytorch/audio}.
Speech features can also be pre-computed and stored in NumPy~\citep{harris2020array} format. Optionally, raw audio files or features files can be packed into ZIP archives to improve I/O performance or facilitate file management. For further pre-processing, \stot~provides online speech data transforms, including CMVN (cepstral mean and variance normalization), speed perturbation~\citep{ko2017study} and SpecAugment~\citep{park2019specaugment}. It also has an open interface for user-defined transforms. For text data, \stot~does online tokenization with a rich collection of tokenizers, including Moses\footnote{https://github.com/moses-smt/mosesdecoder}, SentencePiece~\citep{kudo-richardson-2018-sentencepiece}, subword-nmt\footnote{https://github.com/rsennrich/subword-nmt}, byte-level BPE~\citep{wang2020aaai} and bytes~\citep{li2019bytes}.

\paragraph{Data Configuration} \stot~gets raw audio (feature) paths and target texts from manifest files in TSV (tab-separated values) format, which is similar to Kaldi-style scp files. Online speech data transforms and other data-related settings (e.g. tokenizer type and vocabulary) are defined by a separate configuration file in YAML format.

\paragraph{Computation}
\fairseq~is implemented in PyTorch~\citep{paszke2019pytorch} and it provides efficient batching, mixed precision training~\citep{micikevicius2018mixed}, multi-GPU as well as multi-machine training for computational efficiency on large-scale experiments.

\paragraph{Evaluation Metrics} \stot~provides common automatic metrics for ASR, ST and MT, including WER (word error rate), BLEU~\citep{papineni2002bleu} and chrF~\citep{popovic2015chrf}. It also integrates \textsc{simuleval}~\citep{ma2020simuleval} for simultaneous ST/MT metrics such as AL (average lagging)~\citep{Ma2018STACLST} and DAL (differentiable average Lagging)~\citep{cherry2019thinking}.

\paragraph{Visualization} \fairseq~supports Tensorboard\footnote{https://github.com/tensorflow/tensorboard} for monitoring holistic metrics during model training. It also has VizSeq~\citep{wang-etal-2019-vizseq} integration for sequence-level error analysis, where speech and target/predicted text data are visualized with alignments in Jupyter Notebook interface.

\begin{table*}[t]
    \centering
    \small
    \begin{tabular}{rr|cccccccc}
    \toprule
    & & De & Nl & Es & Fr & It & Pt & Ro & Ru \\
    \midrule
    \multicolumn{2}{r|}{Transformer$^1$} & 17.3 & 18.8 & 20.8 & 26.9 & 16.8 & 20.1 & 16.5 & 10.5 \\
    \multicolumn{2}{r|}{Transformer$^{2\dagger}$} & 22.9 & 27.4 & 28.0 & 32.7 & 23.8 & 28.0 & 21.9 & 15.8  \\
    \midrule
    \multicolumn{2}{r|}{T-Sm} & 22.7 & 27.3 & 27.2 & 32.9 & 22.7 & 28.1 & 21.9 & 15.3 \\
    \multicolumn{2}{r|}{Multi. T-Md$^\ast$} & 24.5 & 28.6 & 28.2 & 34.9 & 24.6 & 31.1 & 23.8 & 16.0 \\
    \midrule
    \multirow{4}{*}{\rotatebox[origin=c]{90}{B-Base}} & Offline & 19.2	&23.5&	24.0&	29.1&	16.4&	23.5&	19.7&	13.7 \\
    & High Lat.$^\ddagger$ & 18.6 (6.8) &  22.9 (6.9) &  22.3 (6.8)& 28.4 (6.7)& 15.4 (6.8)& 22.6 (6.9)&  19.1 (6.7)& 12.9 (6.9)\\
    & Mid Lat.$^\ddagger$ & 14.1 (5.4) &  17.9 (5.4) &  17.2 (5.5)& 25.0 (5.3)& 12.0 (5.5)& 17.7 (5.8)&  15.0 (5.6)& 7.2 (5.8)\\
    & Low Lat.$^\ddagger$
    & 8.2 (2.9)  &  12.3 (2.8) &  13.0 (3.0)& 21.1 (2.8)& 6.7 (2.9)& 13.3 (2.9)&  12.1 (2.9)& 4.9 (2.7)\\
    \bottomrule
    \end{tabular}
    \caption{\stot~models on MuST-C. Test BLEU reported (for online models, AL is shown in parentheses). $^1$~\citet{9004003}. $^2$~\citet{inaguma-etal-2020-espnet}.  $^\dagger$ Applied additional techniques: speed perturbation, pre-trained decoder from MT and auxiliary CTC loss for ASR pre-training. $^\ddagger$ Online models using beam size of 1 (instead of 5). $^\ast$ Trained jointly on all 8 languages.}
    \label{tab:mustc_st_results}
\end{table*}
\begin{table}[t]
    \centering
    \small
    \begin{tabular}{r|ccc}
    \toprule
     &  Type & Config. & Params \\
    \midrule
    B-Base & \multirow{2}{*}{RNN$^1$} & 512d, 3L enc./2L dec. & 31M \\
    B-Big &  & 512d, 5L enc./3L dec. & 52M \\
    \midrule
    T-Sm & Trans- & 256d, 12L enc./6L dec. & 31M \\
    T-Md & former$^2$ & 512d, 12L enc./6L dec. & 72M \\
    T-Lg & & 1024d, 12L enc./6L dec. & 263M \\
    % T-XL & & 1024d, 16L enc./6L dec. & 318M \\
    \midrule
    W-Lg & wav2vec & 1024d, 24L & 315M \\
    CW-Lg & 2.0$^3$ & 1024d, 24L, Conformer$^4$ & 618M \\
    \bottomrule
    \end{tabular}
    \caption{\stot~models for benchmarking. For simplicity, we use the same (default) model hyper-parameters and learning rate schedule across all experiments. $^1$~\citet{berard2018end}.
    $^2$~\citet{vaswani2017attention}. $^3$~\citet{NEURIPS2020_92d1e1eb}. $^4$~\citet{gulati20_interspeech}.
    }
    \label{tab:models}
\end{table}

\begin{table}[t]
    \centering
    \small
    \begin{tabular}{r|cc|cc}
    \toprule
    & \multicolumn{2}{c|}{Dev} & \multicolumn{2}{c}{Test} \\
    & Clean & Other & Clean & Other \\
    \midrule
    \multicolumn{5}{c}{100h labeled} \\
    \midrule
    W-Lg$^3$ & 3.3 & 6.5 & 3.1 & 6.3 \\
    \midrule
    T-Sm & 14.0 & 28.7 & 15.3 & 29.6 \\
    + CTC Aux. & 11.8 & 26.8 & 13.9 & 27.3 \\
    CW-Lg & 2.5 & 5.0 & 2.5 & 5.0 \\
    \midrule
    \multicolumn{5}{c}{960h labeled} \\
    \midrule
    LAS$^1$ & - & - & 2.8 & 6.8 \\
    Transformer$^2$ & 2.5 & 6.7 & 2.9 & 7.0 \\
    W-Lg$^3$ & 2.1 & 4.5 & 2.2 & 4.5 \\
    CW-Lg$^4$ & 1.7 & 3.5 & 1.7 & 3.5 \\
    \midrule
    B-Big & 3.7 & 11.4 & 3.9 & 11.5 \\
    T-Sm & 3.8 & 8.9 & 4.4 & 9.0 \\
    T-Md & 3.2 & 8.0 & 3.4 & 7.9 \\
    T-Lg & 3.0 & 7.5 & 3.2 & 7.5 \\
    CW-Lg & 1.7 & 3.5 & 1.8 & 3.7 \\
    \bottomrule
    \end{tabular}
    \caption{\stot~models on LibriSpeech. Dev and test WER reported. $^1$~\citet{park2019specaugment}. $^2$~\citet{synnaeve2019end}. $^3$~\citet{NEURIPS2020_92d1e1eb}. $^4$~\citet{zhang2020pushing}.}
    \label{tab:librispeech_results}
\end{table}

\begin{table*}[t]
    \small
    \centering
    \begin{tabular}{r|ccccccccccccc}
    \toprule
         & Fr & De & Es & Zh & Tr & Ar & Sv & Lv & Sl & Ta & Ja & Id & Cy \\
         \midrule
         \multicolumn{14}{c}{X$\rightarrow$En} \\
         \midrule
         B-Base & 23.2 & 15.7 & 20.2 & 4.4 & 2.2 & 2.7 & 1.4 & 1.2 & 1.5 & 0.2 & 1.1 & 1.0 & 1.7 \\
         + SSL$^\star$ & 23.1& 16.2& 20.2& 4.8& 3.2& 3.8& 3.7& 2.3& 2.2& 0.2& 1.6& 1.6& 2.2 \\
         Multi. B-Big$^\dagger$ & 26.6 & 19.5 & 26.3 & 4.4 & 2.1 & 0.3 & 1.3 & 0.6 & 1.4 & 0.1 & 0.6 & 0.3 & 0.9 \\
         T-Sm & 26.3 & 17.1 & 23.0 & 5.8 & 3.6 & 4.3 & 2.7 & 2.5 & 3.0 & 0.3 & 1.5 & 2.5 & 2.7 \\
         Multi. T-Md$^\dagger$ & 26.5 & 17.5 & 27.0 & 5.9 & 2.3 & 0.4 & 0.5 & 0.6 & 0.7 & 0.1 & 0.1 & 0.3 & 1.9 \\
         \midrule
         \multicolumn{14}{c}{En$\rightarrow$X} \\
         \midrule
         B-Base & - & 12.5 & - & 20.0 & 6.7 & 9.1 & 18.1 & 8.7 & 11.6 & 7.4 & 25.6 & 15.2 & 18.9 \\
         Multi. B-Big$^\ddagger$ & - & 12.6 & - & 22.2 & 7.3 & 8.0 & 18.3 & 8.9 & 11.4 & 7.3 & 28.2 & 16.0 & 19.3 \\
         T-Sm & - & 16.3 & - & 25.4 & 10.0 & 12.1 & 21.8 & 13.0 & 16.0 & 10.9 & 29.6 & 20.4 & 23.9 \\
         Multi. T-Md$^\ddagger$ & - & 15.4 & - & 26.5 & 9.5 & 10.8 & 20.9 & 12.2 & 14.6 & 10.3 & 30.5 & 18.9 & 22.0 \\
    \bottomrule
    \end{tabular}

    \caption{\stot~models on CoVoST 2. Test BLEU reported (character-level BLEU for Zh and Ja targets). $^\star$ Replaced mel-filter bank features with wav2vec ones ~\citep{Schneider2019,wu2020selfsupervised}. $^\dagger$ Trained jointly on all 21 X-En directions with temperature-based (T=2) resampling~\citep{arivazhagan2019massively}. $^\ddagger$ Trained jointly on all 15 En-X directions.}
    \label{tab:covost2_results}
\end{table*}

\section{Experiments}
We evaluate \stot~models on English ASR benchmark---LibriSpeech~\citep{panayotov2015librispeech}, as well as multilingual ST benchmarks---MuST-C~\citep{di2019must} and CoVoST 2~\citep{wang2020covost}. The model architectures used in benchmarking can be found in Table~\ref{tab:models}.

\subsection{Experimental Setup}
For speech inputs, we extract 80-channel log mel-filter bank features (25ms window size and 10ms shift) with utterance-level CMVN applied. We remove training samples with more than 3,000 frames for GPU memory efficiency. To alleviate overfitting, we pre-train ST model encoders on English ASR and adopt SpecAugment (without time warping): LD policy on LibriSpeech models and LB policy on MuST-C and CoVoST 2 models. We average the last 10 checkpoints and use a beam size of 5 for decoding. For ASR, we use 10K unigram vocabulary~\citep{kudo-richardson-2018-sentencepiece} and report WER. For ST, we use character vocabulary for CoVoST 2 and 8K unigram vocabulary for MuST-C. We report case-sensitive detokenized BLEU using sacreBLEU~\cite{post-2018-call}, except for Japanese and Chinese translations (no word segmentation) where we report character-level BLEU.

\subsection{Speech Recognition (ASR)}
LibriSpeech is a de-facto standard ASR benchmark that contains 1,000 hours of English speech from audiobooks. Table~\ref{tab:librispeech_results} shows the dev and test WER of our models on LibriSpeech clean and noisy sets. Three architectures, RNN-based model (``B-Big"), Transformer-based models (``T-Sm", ``T-Md" and ``T-Lg") and Conformer-based wav2vec 2.0 model (``CW-Lg"), are evaluated. We can see that the first two architectures are able to achieve competitive performance (WER) to the state-of-the-art ones, while we use only default model hyper-parameters and learning rate schedule without any task-specific tuning. Our implementation of the third architecture matches the state of the art.

\subsection{Speech Translation (ST)}

\subsubsection{MuST-C}
MuST-C contains up to around 500 hours of English speech from TED talks with translations in 8 European languages.
Table~\ref{tab:mustc_st_results} shows the test BLEU of our Transformer-based models (``T-Sm" and ``Multi. T-Md") and RNN-based models (``B-Base") on all the MuST-C language directions.
Compared with previous Transformer-based approaches \cite{di-gangi-etal-2019-enhancing, inaguma-etal-2020-espnet}, our bilingual models achieve comparative results to the state of the art without applying additional techniques such as speed perturbation and pre-trained decoder from MT. Moreover, our multilingual model (trained on all 8 languages) outperforms all bilingual ones with large margins. Besides traditional offline models, we also provide simultaneous ST models: the lower section in Table~\ref{tab:mustc_st_results} presents the online models with wait-$k$ policy, which was the baseline system in the IWSLT 2020 shared task on simultaneous ST~\citep{ansari-etal-2020-findings}. 
The results represent the best systems in
high ($\textrm{AL} > 6$), medium ($6 \ge \textrm{AL} > 3$) and low ($\textrm{AL} \le 3$) latency regimes, on which we can clearly see the trade-offs between model performance and prediction latency.

\subsubsection{CoVoST 2}
CoVoST 2 contains total 2,880 hours of read speech in 22 languages from the open-source community, with 21 X-En directions and 15 En-X directions. We evaluate our models bidirectionally on 13 languages of them, including low-resource X-En directions: Zh, Tr, Ar, Sv, Lv, Sl, Ta, Ja, Id and Cy. We observe from Table~\ref{tab:covost2_results} that our Transformer-based models (``T-Sm" and ``T-Md") outperforms RNN-based ones (``B-Base" and ``B-Big") on all En-X and X-En directions. The performance gap tends to be larger when the training data is higher resource (En-X directions, Fr-En, De-En and Es-En). Our multilingual models perform reasonably well with a universal model for over 15 X-En or En-X directions. They even have significant improvements on some directions (e.g. at least 4 BLEU gain on Es-En). For low-resource directions, we also evaluate self-supervised speech features~\citep{Schneider2019,wu2020selfsupervised}\footnote{From a wav2vec model pre-trained on LibriSpeech: \url{https://github.com/pytorch/fairseq/tree/master/examples/wav2vec}} as an alternative to the traditional log mel-filter bank features (``+ SSL"). We find that self-supervised features bring consistent gains and transfer well across different languages (self-supervised model trained on English and feature extracted for non-English).

\section{Conclusion}
We introduce \stot, a \fairseq~extension for speech-to-text (S2T) modeling tasks such as speech recognition and speech translation. It includes end-to-end workflows and state-of-the-art models with scalablity and extensibility design. It seamlessly integrates \fairseq's machine translation models and language models to improve S2T model performance. \stot~documentation and examples are available at \url{https://github.com/pytorch/fairseq/tree/master/examples/speech_to_text}.

\section*{Acknowledgments}
We thank Myle Ott, Michael Auli, Alexei Baevski, Jiatao Gu, Abdelrahman Mohamed and Javad Dousti for helpful discussions.

\bibliography{aacl-ijcnlp2020}
\bibliographystyle{acl_natbib}

\end{document}